# PerceptronCARE: A Deep Learning-Based Intelligent Teleophthalmology Application for Diabetic Retinopathy Diagnosis

Akwasi Asare[1*], Isaac Baffour Senkyire[1], Emmanuel Freeman[1], Mary Sagoe[1], Simon Hilary Ayinedenaba Aluze-Ele[2], and Kelvin Kwao[3]

[1]Computer Science Department, Ghana Communication Technology University, Accra, PMB 100, Ghana

[2]Computer Science Department, C K Tedam University, Navrongo, Box 24, Ghana

[3]Public Utilities Regulatory Commission, Accra, Ghana

[*]Corresponding Author: Akwasi Asare. Email: nasare34@yahoo.com

**Abstract:** Diabetic retinopathy is a leading cause of vision loss among adults and a major global health challenge, particularly in underserved regions. This study presents PerceptronCARE, a deep learning-based teleophthalmology application designed for automated diabetic retinopathy detection using retinal images. The system was developed and evaluated using multiple convolutional neural networks, including ResNet-18, EfficientNet-B0, and SqueezeNet, to determine the optimal balance between accuracy and computational efficiency. The final model classifies disease severity with an accuracy of 85.4%, enabling real-time screening in clinical and telemedicine settings. PerceptronCARE integrates cloud-based scalability, secure patient data management, and a multi-user framework, facilitating early diagnosis, improving doctor-patient interactions, and reducing healthcare costs. This study highlights the potential of AI-driven telemedicine solutions in expanding access to diabetic retinopathy screening, particularly in remote and resource-constrained environments.

# 1 INTRODUCTION

Diabetes is a significant global health challenge, and its rates have risen worldwide [1]. It is a major cause of death and disability, affecting people regardless of country, age group, or sex[2]. Diabetic Retinopathy (DR) is one of the major complications of diabetics and represents a chronic eye condition that commonly affects individuals with diabetes, often leading to vision loss in the working-age population [1]. The condition is caused by damage to the blood vessels in the retina, which can become swollen and leaky[3]. Approximately 34.6% of people with diabetes develop diabetic retinopathy (DR), the leading cause of vision loss in adults of working age [4]. By 2045, there are expected to be 242 million cases of DR and 71 million instances of advanced DR [4]. Diabetic retinopathy has various stages, from moderate to severe see Fig 1 for DR Stages, and early detection and treatment are essential to prevent vision loss[3] and reduce blindness by 95% [5], [6]. Ophthalmologists use a conventional approach of manually examining digital retinal fundus images, a process which is time-consuming. Hence, Telemedicine a time and cost-saving technology that has been introduced in the detection of DR, which improves hospital and clinic operations by monitoring discharged patients and aiding in their recovery [7]. The progression of Diabetic Retinopathy (DR) through its various stages is illustrated in **Figure 1**, providing visual context for the severity levels considered in this study.

Telemedicine applications such as computer-aided detection (CAD) are instrumental in identifying abnormalities in medical imaging, including chest X-rays and Computed Tomography (CT) scans [8]. CAD leverages deep learning to address this challenge, highlighting its importance in telemedicine [8]. Deep learning is an exciting branch of machine learning that harnesses the power of artificial neural networks to uncover intricate patterns and features from data. By diving deeper into the layers of information, it enables us to capture more complex and abstract characteristics, transforming raw input into valuable insights [9]. Deep learning has successful applications in image identification, voice recognition, and natural language processing, and is increasingly used in medical image reading to assist doctors and improve efficiency [9]. Deep learning also has the potential to transform emergency response, diagnose illnesses, expedite drug discovery, and provide personalized treatments [10], [11], [12]. The rapid rise of telemedicine, fueled by advancements in mobile technology, has fundamentally changed the delivery of medical care [7].

This study focuses on diabetic retinopathy, a major global healthcare challenge. It proposes a Deep Learning-based intelligent teleophthalmology application perceptronCARE to help diagnose and treat the disease early. The application enhances doctor-patient communication, reduces healthcare costs, and increases patient involvement. The goal is to improve treatment outcomes and reduce mortality related to diabetic retinopathy.

The study is organized into several sections: (1) Introduction, which provides an overview of diabetic retinopathy and the role of AI in teleophthalmology; (2) Health Applications, discussing the impact of AI-driven solutions in healthcare; (3) Diabetic Retinopathy Application, explaining the condition, its progression, and the need for early detection; (4) Literature Review of Diabetic Retinopathy, summarizing existing research and identifying gaps; (5) Methodology, detailing dataset preprocessing, model selection, training, and evaluation; (6) Results and Discussion, presenting model performance, comparative analysis, and system scalability; (7) Presenting Our Application: PerceptronCARE Diabetic Retinopathy Application, describing its functionalities and user roles; and (8) Conclusion and Future Work, summarizing key findings and outlining directions for improvement. These sections offer a comprehensive overview of diabetic retinopathy, leveraging advanced technologies like Deep learning to improve the accuracy and speed of medical examinations using the proposed application.

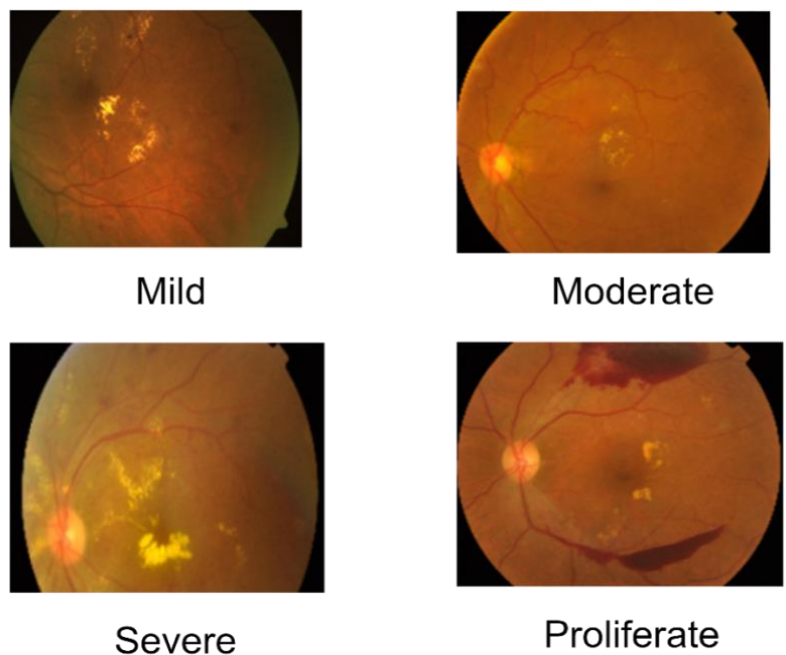

**Figure 1**: DR Stages *([13])*

## 2 HEALTH APPLICATIONS

The field of health research has greatly broadened, ranging from topics like brain hemorrhage[14] obesity[15], diabetes [16], heart disease including heart failures[17], [18], gastrointestinal diseases [19], and chronic illness and older adult care [20]. "For people with diabetes, especially type 2 diabetes, the use of a mobile health (mHealth) app has been associated with a moderate improvement in glycemic control (reduction of 0.2%-0.4%—2-4 mmol/mol—in HbA1c)" according to [21]. These medical disciplines, which concentrate on enhanced diagnostic services, illness screening, prevention, and surveillance, are mostly used in electronic health (e-health)[22]. **Table 1** presents a list of some of the most widely used health applications, categorized by their origin, developing company, and application type.

**Table 1:** Existing Health Applications

| Application Name | Origin | Company | Application Type | Reference |
|---|---|---|---|---|
| MyFitnessPal | USA | Under Armour | Health & Fitness | [23] |
| Medisafe | USA | Medisafe Project | Medication | [24] |
| Glucose Buddy | USA | Glucose Buddy | Diabetes | [25] |
| CareZone | USA | CareZone Inc. | Health Management | [26] |
| Clue | Germany | Clue GmbH | Menstrual Tracking | [23] |
| Sleep Cycle | Sweden | Northcube AB | Sleep Tracking | [24] |
| Zocdoc | USA | Zocdoc, Inc. | Telemedicine | [25] |
| Doctor on Demand | USA | Doctor on Demand, Inc. | Telemedicine | [26] |

Maaß and colleagues describe health applications as mobile applications that manage health-related information for or about users [27]. Health Applications are designed for those who want to maintain, improve, or monitor their own or a community's health. Medical apps come under this category of health applications since they provide comparable technical capabilities and services. According to [28], users of m-health applications are classified as active, or passive based on how they interact with the apps. Active users fully make use of app features, resulting in advantages such as better mental well-being, whereas passive users engage seldom with restricted functions. Active participation in m-health leads to improved health outcomes and cost savings. The major target users are healthcare professionals, patients, and family caregivers. Medical apps are intended for clinical and medical purposes and can be legally regulated as mobile medical devices [27]. In the last ten years, the mobile health (m-health) app market has developed and released hundreds of thousands of applications each year. These applications vary widely, encompassing those created by prominent high-tech IT companies that target a global market. As a result, most mobile users now have at least one m-health app downloaded on their devices [28], [29]. For example, the COVID-19 pandemic has accelerated m-health adoption, setting the path for future industry development across multiple user groups.

M-health applications are divided into five categories: clinical, registration/guidance, pharmaceutical, diet/health monitoring, and patient support [28]. Mobile health apps are crucial because they address the significant challenges underdeveloped nations face, where health facilities are often far away, unreachable, or nonexistent. For example, the World Health Organization (WHO) promotes diabetic retinopathy (DR) screening operations, but widespread implementation requires substantial investments in manpower, supplies, and financial backing, which many low- and middle-income countries struggle to meet [4]. This shows that there is still a gap between technological discoveries and their practical adoption in clinical settings. The suboptimal treatment of this DR underscores the urgent need for technology to improve diagnosis and treatment outcomes. This is why mobile applications for health systems are needed and increasingly necessary, being developed, and undergoing significant advancements. Mobile health systems often include telephony, data management, and message services (such as email, SMS, and MMS) [22]. The increasing usage of mobile devices throughout the world provides the potential to enhance healthcare access and response to health crises. It is critical for health authorities, particularly in developing countries, to aggressively exploit the benefits of mobile technology in the healthcare business [22].

## 3 DIABETIC RETINOPATHY APPLICATION

Diabetic retinopathy is a serious eye condition caused by diabetes that can lead to vision loss if untreated [3], [5]. Health diagnostic technologies have advanced significantly, with computer-aided diagnosis (CAD) being a major breakthrough. CAD systems, used globally in hospitals and clinics, employ Convolutional Neural Networks (CNNs) for image classification [30]. Deep learning methods, especially CNNs, show promise in automating the detection and classification of DR, offering precise feature extraction and accurate severity predictions.

CNNs consist of three main layers: Convolution (CONV), Pooling, and Fully Connected (FC), each playing a crucial role in processing and analyzing medical images [10], [31].

Key applications of deep learning in DR involve:

- Identifying and classifying different phases of DR automatically, including neovascularization, microaneurysms, hemorrhages, exudates, cotton wool patches, and macular edema [10].
- Adjusting pretrained deep learning models for DR diagnosis by transfer learning techniques, improving performance under constrained training scenarios. Employing ensemble methods that combine many deep learning models to increase the classification accuracy of diabetic retinopathy, and
- Using preprocessing methods such as contrast-limited adaptive histogram equalization (CLAHE) to enhance lesion visibility in retinal images, thereby increasing the effectiveness of deep learning models [32].

In this study, a pretrained CNN model named ResNet-18 was utilized. It was trained on the ImageNet dataset through transfer learning, embedded into the perceptronCare application to detect severity levels in fundus images for diagnosing DR.

## 4 LITERATURE REVIEW OF DIABETIC RETINOPATHY

Diabetic Retinopathy (DR) stands as the leading cause of visual impairment and blindness among individuals of working age globally [33]. This condition, stemming from prolonged diabetes affecting retinal blood vessels, results in complications such as macular edema and retinal neovascularization, ultimately leading to vision impairment and blindness [34], [35]. The prevalence of DR increases with the duration of diabetes, affecting approximately one-third of diabetic individuals [33]. Early identification and treatment of DR can significantly reduce the likelihood of severe vision loss by up to 95% [5], [6]. Further studies should focus on the barriers to early screening and treatment adherence to improve these outcomes.

Deep learning techniques have shown promise in automating the identification and classification of diabetic retinopathy[36]. Recent advancements, particularly convolutional neural networks (CNNs), have revolutionized medical image analysis by enabling precise feature extraction and accurate severity predictions [10], [31]. CNNs excel in identifying complex patterns and structures within medical images, facilitating the detection of abnormalities, including cancers, and the segmentation of organs [33]. Despite these advancements, there is still a need for standardized evaluation metrics to compare the performance of different CNN models effectively.

The integration of deep learning into telemedicine such as teleophthalmology, a digital technology, has further enhanced the screening and management of DR. This integration represents a significant shift towards more accessible and efficient tools for diabetes care [37]. The integration of deep learning into telemedicine, such as teleophthalmology, a digital technology, has further enhanced the screening and management of DR. This integration represents a significant shift towards more accessible and efficient tools for diabetes care [37]. However, the study by [37] does not address the challenge of adapting these technologies to diverse and underserved populations, which is crucial for ensuring equitable access to DR screening [37]. Our study aims to address this by providing comprehensive data analysis and monitoring capabilities that track long-term patient outcomes and cost efficiency. PerceptronCARE is relevant here as it is designed to be adaptable and scalable, potentially providing tailored solutions for various populations and settings, thus addressing this gap in accessibility.

Digital technologies, especially smartphone-based deep learning applications, have been instrumental in the screening and management of DR. Studies have demonstrated the efficacy of CNNs in DR detection across diverse datasets such as DIARETDB0, DRIVE, CHASE, and the Kaggle Dataset. These models not only enhance diagnostic accuracy but also improve efficiency and reduce costs compared to traditional methods [34]. Despite these advancements, there is limited research on the real-world deployment and user experience of these technologies in everyday clinical practice. Our study aims to bridge this gap by integrating user-friendly interfaces and robust performance in real-world conditions, ensuring that the technology is practical and effective in diverse clinical environments. PerceptronCARE offers a scalable solution designed to integrate seamlessly into existing healthcare infrastructures, ensuring widespread adoption and consistent performance across various clinical environments.

Researchers have explored various approaches to DR detection using deep learning. For instance, [38] investigated diabetic retinopathy across five severity levels using fundus images without requiring any pre- or post-processing. Their approach, incorporating fine-tuning stages and a semi-supervised deep learning technique, achieved sensitivity of 92.18% and specificity of 94.50%. However, this study does not address the limitations related to the generalizability of their model to different populations and variations in imaging quality. As a result, a validation across larger and more diverse populations is needed to confirm these findings. Our study aims to address this shortcoming by employing advanced algorithms and extensive training datasets to improve model generalizability and robustness across various demographic and imaging conditions. PerceptronCARE's robust validation framework includes extensive testing across multiple demographics and data sets, ensuring generalizability and reliability.

Using the Kaggle Dataset, a custom convolutional neural network (CNN) model was created to categorize retinal pictures into five severity categories of DR; preprocessing procedures included color normalization and picture scaling to 226 × 226 pixels. In order to reduce overfitting, regularization strategies such as dropout and L2 regularization were used. This led to metrics that were attained with 95% specificity, 75% accuracy, and 30% sensitivity [31]. These results highlight the need for further optimization to improve sensitivity thus further refinement is needed to detect DR at earlier stages. PerceptronCARE incorporates advanced regularization techniques and a novel architecture designed to enhance sensitivity while maintaining high specificity.

Smartphone-based AI programs have shown exceptional diagnostic accuracy in identifying DR and referable DR, with combined sensitivity ranging from 89.5% to 97.9% and specificity from 85.9% to 92.4% [39], [40]. Future research should focus on the user-friendliness and accessibility of these programs to ensure widespread adoption. While these programs are highly accurate, they often lack integration with comprehensive patient management systems, which can limit their effectiveness in providing holistic care. Our application prioritizes user-centered design, offering an intuitive interface and comprehensive support to facilitate adoption by healthcare providers and patients alike. Also, our application enhances this by integrating diagnostic capabilities with patient management tools, thus offering a more complete solution for DR care and follow-up.

Notably, the ResNet-152 model achieved an outstanding accuracy of 99.41% in detecting DR using the Kaggle EyePACS Dataset[41] . These promising results warrant further investigation into the model's performance across different demographic groups and varying image qualities; the model's performance was evaluated using a single dataset, which may not fully represent variability in clinical practice. PerceptronCARE incorporates dataset from the EyePACS (Kaggle) dataset which was curated from different racial groups and settings in its training pipeline to ensure robust performance across varied demographics and image conditions, addressing this gap effectively.

Moreover, the same Kaggle Dataset was used to assess pretrained CNN models (VGG16, InceptionNet V3, and AlexNet), which produced average accuracies of 50.03% for VGG16, 63.23% for InceptionNet V3, and 37.43% for AlexNet. The authors explained these outcomes by pointing out that there were only 166 photos in the training set [31]. Future work should include larger and more balanced datasets to enhance model training and performance. Our study leverages extensive and balanced datasets of 3667, ensuring robust training and high-performance metrics.

Conversely, [42] used a smartphone linked to a hand-held ophthalmoscope to directly collect retinal fundus pictures, they then developed a pattern recognition classifier utilizing four image sets from the MESSIDOR Database. Their technique has an average sensitivity of 86% and an AUC of 0.844. Additional research is needed to explore the integration of such methods into routine clinical practice and their impact on patient outcomes. PerceptronCARE integrates seamlessly with portable diagnostic devices, enhancing usability in routine clinical practice and aiming to impact patient outcomes significantly.

## 5 METHODOLOGY

The PerceptronCARE system is designed as a robust, real-time telemedicine platform for diabetic retinopathy (DR) screening, integrating deep learning models to ensure high diagnostic accuracy, computational efficiency, and scalability. The methodology follows a structured approach encompassing dataset preparation, model selection, training and optimization, evaluation, and system integration. The deep learning models were carefully chosen based on their ability to balance accuracy with computational efficiency, making them ideal for deployment in both cloud-based hospital settings and resource-constrained environments such as mobile devices or edge computing platforms. Following rigorous experimentation, ResNet-18, EfficientNet-B0, and SqueezeNet were selected as the most effective architectures. The best-performing model was quantized into 8-bit integer and integrated into the PerceptronCARE system, optimizing real-time inferencing capabilities while maintaining high diagnostic precision.

### *5.1 Dataset Description, Pre-processing and Augumentation*

The dataset used for training the deep learning models in this study was a collection of retinal fundus images provided by the EyePACS organization, hosted on Kaggle [13]. This dataset offers a comprehensive range of retinal images, capturing different stages of diabetic retinopathy (DR). For our analysis, we selected a subset of 3,662 images from the total 35,126 images in the dataset and classified them into five categories based on DR severity: No DR, Mild, Moderate, Severe, and Proliferative DR. Specifically, the dataset comprised 1,805 images with No DR, 370 images with Mild DR, 999 images with Moderate DR, 193 images with Severe DR, and 295 images with Proliferative DR. Sample images from this dataset are presented in the **figure 2.**

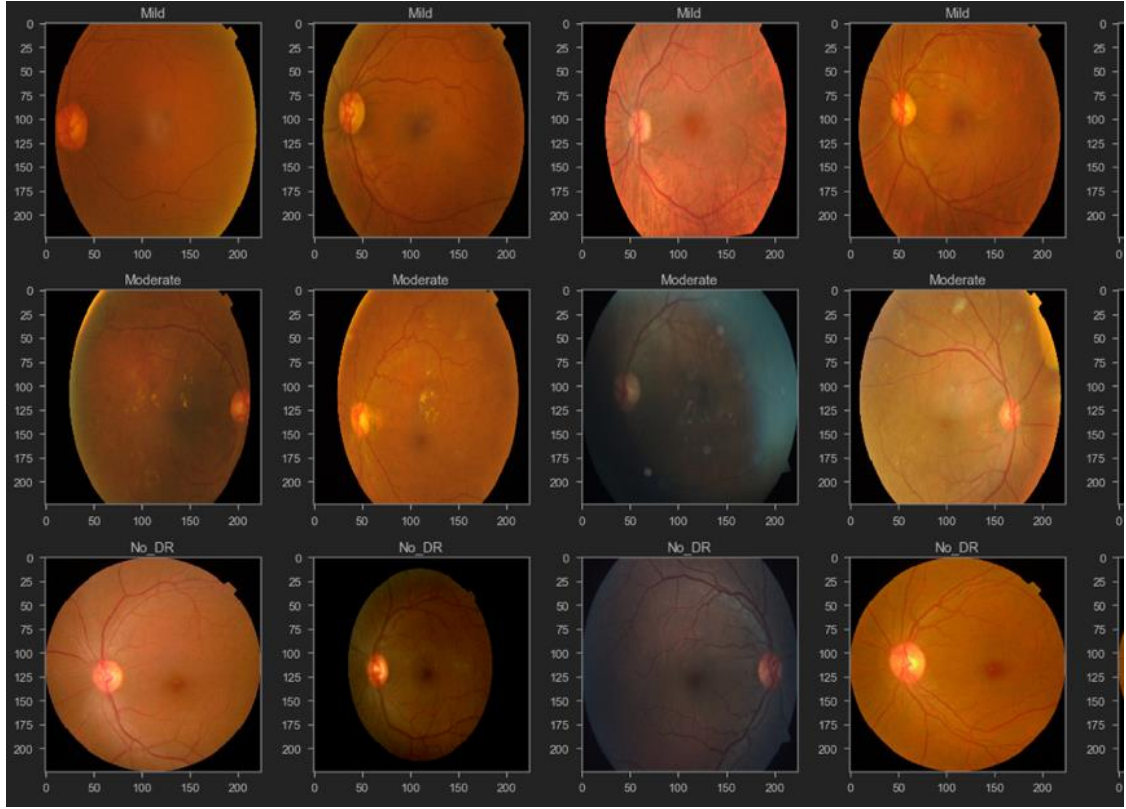

**Figure 2:** Samples of the dataset images.

The PerceptronCARE system is trained and evaluated on the EyePACS dataset, a well-established repository of high-resolution retinal fundus images labeled according to diabetic retinopathy (DR) severity. This selection provided a diverse distribution across different severity levels, which was crucial for training reliable deep-learning models. The dataset comprises five distinct classes: No DR, Mild DR, Moderate DR, Severe DR, and Proliferative DR.

To ensure a representative and unbiased training process, the dataset was split into 70% training, 15% validation, and 15% testing, using stratified sampling to maintain proportional class distribution. However, an analysis of the dataset revealed a significant class imbalance, particularly in the Severe DR and Proliferative DR categories, which contained substantially fewer samples compared to No DR and Mild DR. Such imbalances pose a risk of biased model predictions, where the network might favor more prevalent classes while reducing sensitivity to rare but clinically significant DR stages.

To address this, a targeted data augmentation strategy was employed to increase representation of underrepresented classes while preserving diagnostic features critical for classification. Given that the No DR class was already well-represented, no augmentation was applied to it. However, Mild DR was increased from 370 to 900 images using flipping, rotation, and brightness adjustments to simulate variations observed in real-world retinal scans. Moderate DR, initially comprising 999 images, was expanded to 1,200 to maintain class proportionality. The most extensive augmentation was applied to the Severe DR and Proliferative DR classes, which had the lowest representation. Severe DR was increased from 193 to 900 images using contrast enhancement, noise injection, and zooming to generate synthetic variations while maintaining clinically relevant features. Similarly, Proliferative DR was expanded from 295 to 1,000 images using contrast enhancement and Gaussian noise injection to improve the model's ability to detect this critical stage.

In addition to augmentation, several preprocessing techniques were employed to enhance image quality and optimize model learning. Each image was resized to 226 × 226 pixels to ensure compatibility with the selected architectures. Contrast-limited adaptive histogram equalization (CLAHE) was applied to enhance vessel and lesion visibility, helping the models detect subtle pathological features. Furthermore, pixel normalization was used to scale intensity values between 0 and 1, ensuring a stable training process and improving convergence.

By integrating stratified sampling, augmentation, and preprocessing, the PerceptronCARE system ensures a balanced and effective training process, improving model generalization across all DR severity levels. These modifications enhance the model's sensitivity in detecting severe DR stages, minimizing biases while ensuring reliable and scalable real-world deployment in cloud-based hospital networks and mobile health applications.

To address the issue of class imbalance in the Diabetic Retinopathy (DR) dataset, data augmentation techniques were applied to increase the representation of minority classes. As illustrated in **Figure 3 below**, the dataset originally exhibited a skewed distribution, with a significantly higher number of "No DR" images compared to other severity levels. After augmentation, a more balanced dataset was achieved, particularly enhancing the sample sizes for Mild, Moderate, Severe, and Proliferative DR categories. This balance is crucial for training robust deep learning models and mitigating bias toward the majority class.

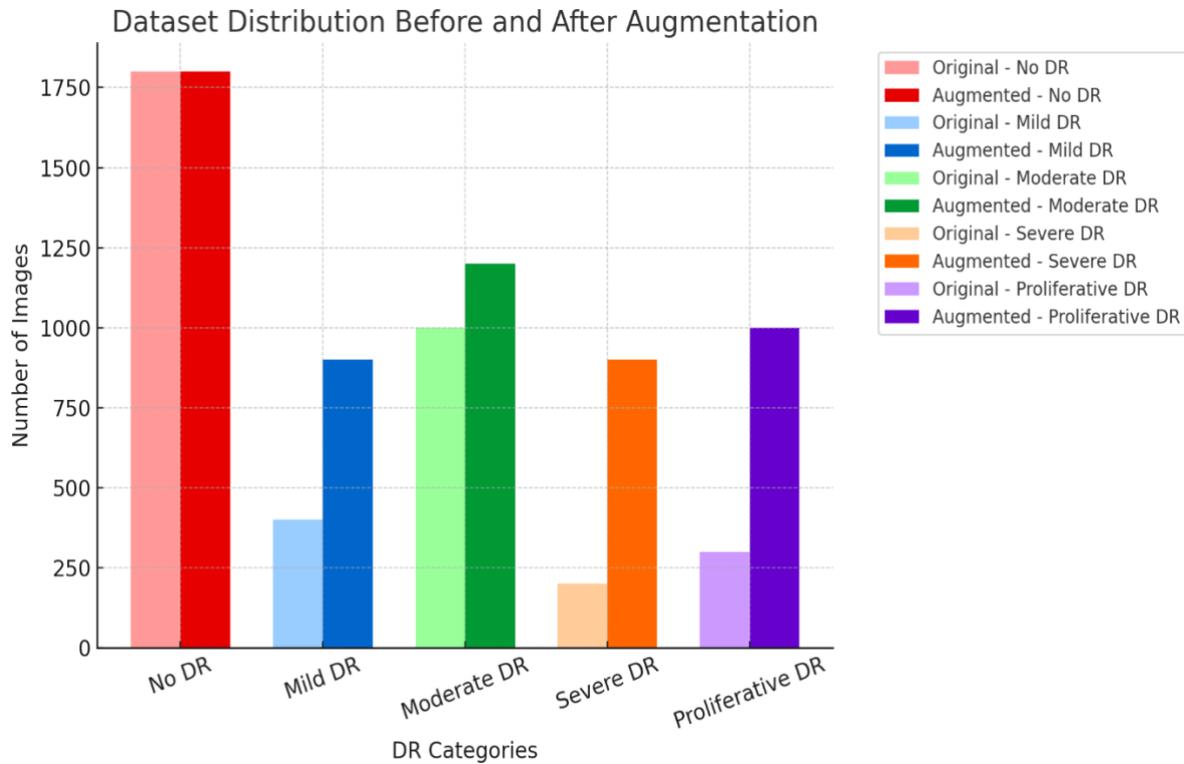

**Figure 3:** Distribution of Diabetic Retinopathy disease dataset showing the original (faded colors) and augmented (non-faded colors) image counts across all DR categories.

The **Table 2** presents the distribution of images across the five diabetic retinopathy (DR) severity levels before and after applying augmentation techniques. The augmentation process was designed to address class imbalance and enhance model generalization, particularly for underrepresented categories.

Table 2: Dataset Breakdown Before and After Augmentation

| DR Category | Original Count | After Augmentation |
|---|---|---|
| **No DR** | 1,805 | 1,805 |
| **Mild DR** | 370 | 900 |
| **Moderate DR** | 999 | 1,200 |
| **Severe DR** | 193 | 900 |
| **Proliferative DR** | 295 | 1,000 |
| **Total** | **3,662** | **5,805** |

## 5.2 Selected Models for Experiment and Architecture Integration into PerceptronCARE

To ensure optimal real-time performance, the PerceptronCARE system leverages three CNN architectures; ResNet-18, EfficientNet-B0, and SqueezeNet each offering unique advantages in feature extraction, model efficiency, and deployment feasibility. The goal is to achieve a seamless balance between diagnostic accuracy and computational efficiency, enabling deployment in diverse healthcare settings. Following the evaluation of all three models, the best-performing model was quantized and integrated into PerceptronCARE, ensuring optimized real-time inference for telemedicine applications. Model quantization significantly reduces memory footprint and latency, making the system adaptable for cloud, mobile, and embedded healthcare devices.

**ResNet-18**, a residual network with 18 layers, is selected for the experiment and would be integrated and tested with the perceptronCare system. Its selection is due to its strong feature extraction capabilities and ability to prevent vanishing gradients via skip connections [43]. It achieves high accuracy while maintaining a relatively low parameter count, making it feasible for both cloud-based and mobile deployments. Its deep residual learning framework enables the model to capture intricate retinal features critical for diabetic retinopathy classification. The architecture of ResNet-18 is illustrated in **Figure 4**, highlighting the key residual blocks and skip connections that enable effective training of deeper networks.

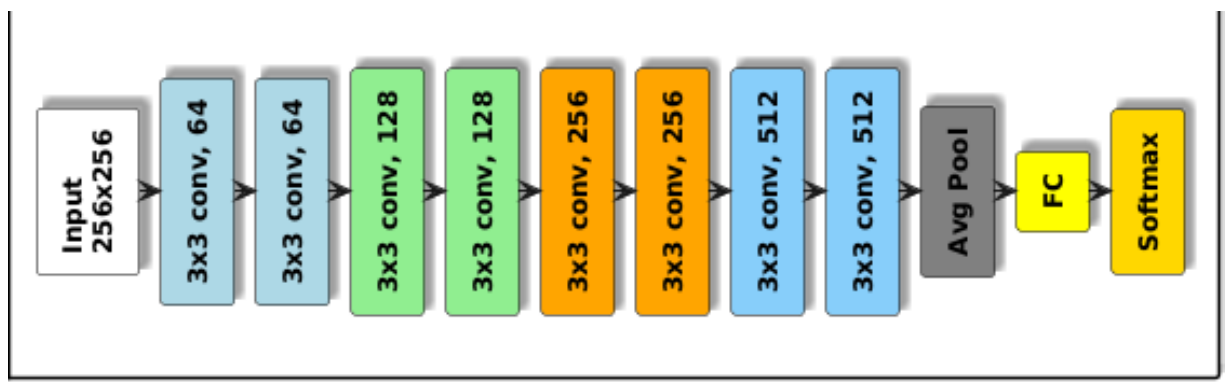

**Figure 4**: ResNet-18 architecture showing the residual learning framework with skip connections for deep feature extraction.

**EfficientNet-B0** is selected for the experiment and would be integrated and tested independently with the PerceptronCARE due to its compound scaling mechanism, which optimizes model depth, width, and resolution for improved classification accuracy with minimal computational cost [44]. Unlike traditional architectures, EfficientNet-B0 is designed to achieve state-of-the-art performance with fewer parameters and lower FLOPs, making it highly efficient for edge computing applications and mobile integration. The structure of EfficientNet-B0 is presented in **Figure 5**, emphasizing its balanced scaling approach and lightweight design.

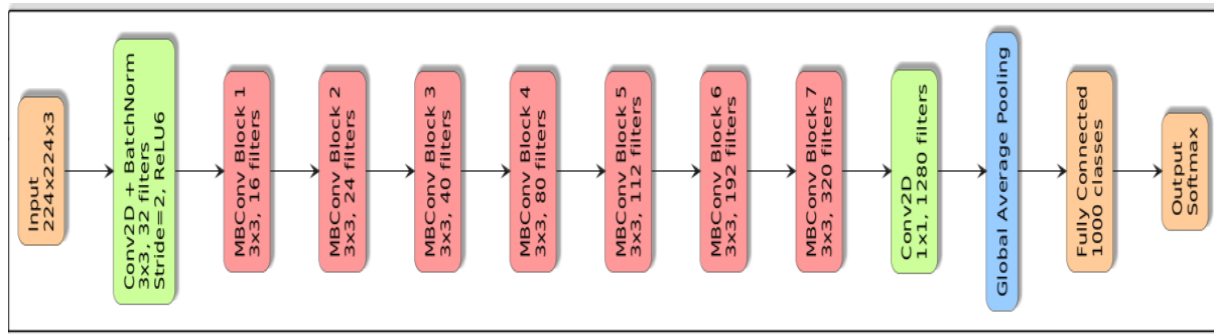

**Figure 5**: EfficientNet-B0 architecture demonstrating compound scaling across depth, width, and resolution for computational efficiency.

**SqueezeNet** was also selected for the experiment and would be integrated and tested with the PerceptronCare system**.** SqueezeNet is known for its ultra-lightweight architecture, designed to drastically reduce model size while maintaining strong classification performance[45]. Through the use of Fire modules, SqueezeNet replaces conventional convolutional layers with squeeze-and-expand operations, allowing it to achieve faster inference times with significantly fewer parameters [45], [46]. This makes it highly suitable for real-time screening in resource-limited healthcare environments where computational resources are constrained. **Figure 6** shows the SqueezeNet architecture, highlighting its compact structure and the role of Fire modules in achieving model compression.

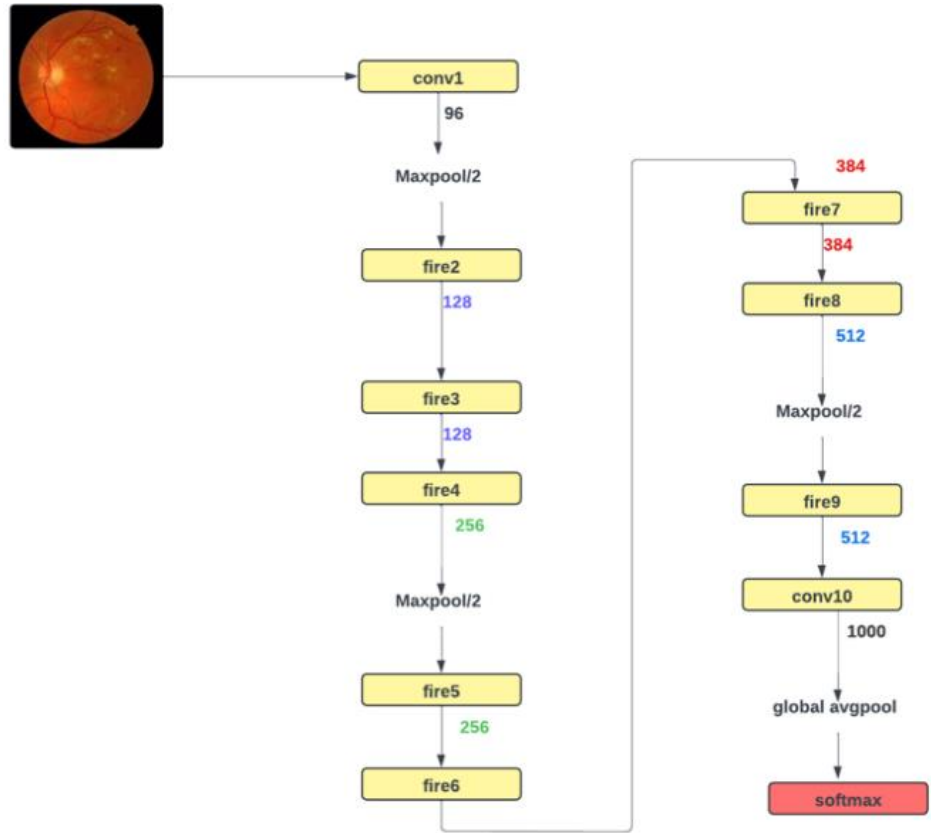

**Figure 6:** SqueezeNet architecture illustrating the use of Fire modules for efficient parameter reduction and fast inference.

The justification for the well performed model based on the experiment would be discussed in the section 6.2.

## 5.3 Model Training and Optimization for PerceptronCARE

The deep learning models were trained using a transfer learning approach, leveraging pre-trained ImageNet weights to accelerate convergence and enhance feature extraction. The final classification layers were customized for the five-class DR classification task to align with PerceptronCARE's clinical requirements. Training was conducted for twenty epochs using the Adam optimizer, with an initial learning rate of 0.0001, which was progressively reduced using a learning rate scheduler to facilitate stable convergence. A batch size of 32 was used, and an early stopping mechanism was implemented to prevent overfitting, halting training once validation loss plateaued. To further improve model robustness, 5-fold cross-validation was employed, allowing performance evaluation across multiple data splits to enhance generalization. The training was conducted on high-performance GPU hardware, optimizing computational efficiency and ensuring real-time diagnostic capabilities for PerceptronCARE's integration.

### 5.4 Performance Evaluation and Deployment Feasibility

To evaluate the performance of the deep learning models integrated into PerceptronCARE, we implemented a comprehensive framework focusing on both diagnostic effectiveness and computational feasibility. This framework was designed to reflect the system's suitability for real-time, cloud-based, and mobile teleophthalmology applications.

The evaluation considered multiple dimensions, including classification accuracy, robustness under class imbalance, inference speed, and resource efficiency. These aspects were crucial to ensure not only clinical reliability but also deployment viability in diverse operational settings, especially those with limited computational infrastructure.

A detailed explanation of the specific evaluation metrics used including accuracy, precision, recall (sensitivity), specificity, F1-score, inference time, number of parameters, FLOPs, and model size—is provided in **Section 5.4.1**.

#### 5.4.1 Evaluation Metrics

Accuracy measures the percentage of images that the model classifies correctly. It is calculated by determining the proportion of true results (both true positives and true negatives) out of all evaluated cases. This metric offers a general overview of how frequently the model's predictions align with the actual classifications.

$$Accuracy = \frac{TP+TN}{TP+TN+FP+FN} \quad (1)$$

Sensitivity, also known as Recall, evaluates how well the model can pinpoint positive cases. It is calculated by taking the ratio of true positives (TP) to the total of true positives and false negatives (FN). This metric underscores the model's capability in identifying occurrences of diabetic retinopathy.

$$Recall = \frac{TP}{TP+FN} \quad (2)$$

Specificity measures the effectiveness of the model in recognizing negative cases. It is determined by the ratio of true negatives (TN) to the total of true negatives and false positives

(FP). This metric indicates the model's ability to accurately identify images that do not show signs of diabetic retinopathy.

$$Specificity = \frac{TN}{TN+FP} \tag{3}$$

Precision, is defined as the proportion of true positives to the overall count of instances classified as positive. It is determined by dividing the count of true positives by the total of true positives and false positives (FP). This metric is essential for evaluating the correctness of the model's positive predictions.

$$Precision = \frac{TP}{TP+FP} \tag{4}$$

The F1 Score, integrates Precision and Recall into one comprehensive metric, making it especially helpful for datasets with imbalanced class distributions. It represents the harmonic mean of Precision and Recall, offering a balanced evaluation that reflects both the model's capacity to identify positive cases and its correctness in positive predictions.

$$F1score = 2 \times \frac{Precision \times Recall}{Precision+Recall} \tag{5}$$

## 6. RESULTS AND DISCUSSION

This section presents a detailed evaluation of the deep learning models tested for diabetic retinopathy classification within PerceptronCARE, a web-based telemedicine system. The primary objective of this evaluation was to identify the best-performing model in terms of classification accuracy, computational efficiency, and real-time feasibility for seamless integration into the system. Three convolutional neural networks (CNNs) were evaluated: ResNet-18, EfficientNet-B0, and SqueezeNet. Each model underwent extensive training and testing, with key metrics such as training and validation accuracy, loss trends, inference time, model size, FLOPs, and AUC scores used to compare their effectiveness. Following experimentation, ResNet-18 emerged as the best-performing model and was the only one integrated into PerceptronCARE, while EfficientNet-B0 and SqueezeNet were retained for comparative analysis but not deployed.

### *6.1 Model Results*

#### 6.1.1 ResNet-18

**Performance and Generalization Ability**

ResNet-18 demonstrated the highest classification accuracy, making it the default model for PerceptronCARE. Its residual learning framework provided effective gradient flow, enabling deeper feature extraction and reducing performance degradation. During training, ResNet-18 achieved a training accuracy of 86.3% with a training loss of 0.211, indicating strong convergence. The validation accuracy was 85.4%, with a validation loss of 0.297, reflecting good generalization capability. The testing accuracy remained consistent at 85.4%, confirming its robustness across unseen data.

**Optimization for Real-Time Deployment**

To ensure efficient real-time inference, ResNet-18 was quantized into 8-bit integer before integration into PerceptronCARE. This significantly reduced its model size and inference latency, allowing for faster processing in telemedicine environments. The quantized ResNet-18 retained its accuracy while reducing computational overhead, making it feasible for both cloud-based hospital networks and mobile health applications.

**Table 3** presents the performance metrics of the quantized ResNet-18 model, which was selected as the core architecture for the PerceptronCARE system. The table highlights the model's high accuracy, efficient inference time, and reduced computational demands.

Table 3: ResNet-18 Performance

| Metric | Value |
| --- | --- |
| Training Accuracy | 86.3% |
| Validation Accuracy | 85.4% |
| Testing Accuracy | 85.4% |
| Training Loss | 0.211 |
| Validation Loss | 0.297 |
| Testing Loss | 0.310 |
| Model Size (Quantized) | 8.18 MB |
| Inference Time | 18 ms |
| FLOPs | 1.8 B |
| AUC Score | ~0.97 |

The training behavior of ResNet-18 is further analyzed to understand its learning efficiency and generalization capability. **Figure 7a** illustrates the training and validation accuracy and loss trends for ResNet-18, demonstrating stable convergence and minimal overfitting across epochs. This consistent performance reinforces the model's suitability for integration into the PerceptronCARE system.

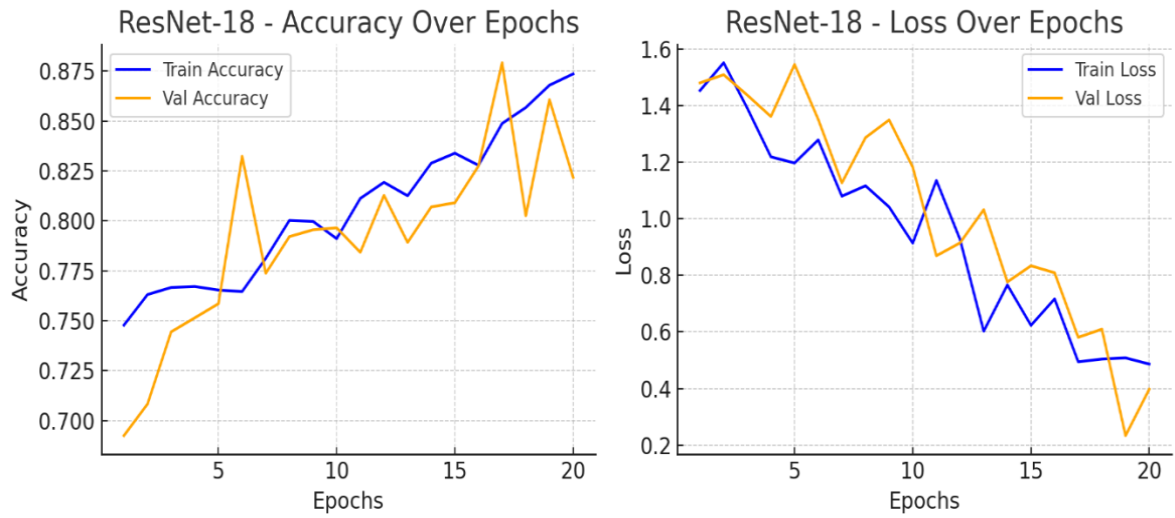

**Figure 7a**: Training and validation accuracy and loss trends for ResNet-18, indicating effective learning and generalization across epochs.

### 6.1.2 EfficientNet-B0

**Trade-offs Between Accuracy and Computational Efficiency**

EfficientNet-B0 was tested due to its optimized compound scaling, which allows for a balanced trade-off between accuracy and efficiency. The model achieved a training accuracy of 84.7% and a validation accuracy of 84.3%, demonstrating strong generalization ability but slightly lower classification accuracy than ResNet-18. EfficientNet-B0's primary advantage was its low number of FLOPs, making it computationally efficient for mobile and edge deployment scenarios. However, its slightly lower recall and precision scores suggested potential misclassification of DR cases, making it less suitable for PerceptronCARE's clinical application. **Table 4** outlines the evaluation results of EfficientNet-B0, showcasing its balance between accuracy and computational efficiency.

Table 4: EfficientNet-B0 Performance

| Metric | Value |
|---|---|
| Training Accuracy | 84.7% |
| Validation Accuracy | 84.3% |
| Testing Accuracy | 84.1% |
| Training Loss | 0.312 |
| Validation Loss | 0.358 |
| Testing Loss | 0.372 |
| Model Size | 9.67 MB |
| Inference Time | 14 ms |
| FLOPs | 0.9 B |
| AUC Score | ~0.95 |

**Figure 7b** presents the training and validation performance of EfficientNet-B0, showcasing its ability to learn efficiently with minimal fluctuations. The model achieves steady improvements in accuracy while maintaining low training and validation losses, highlighting its potential for resource-efficient deployment in the PerceptronCARE system.

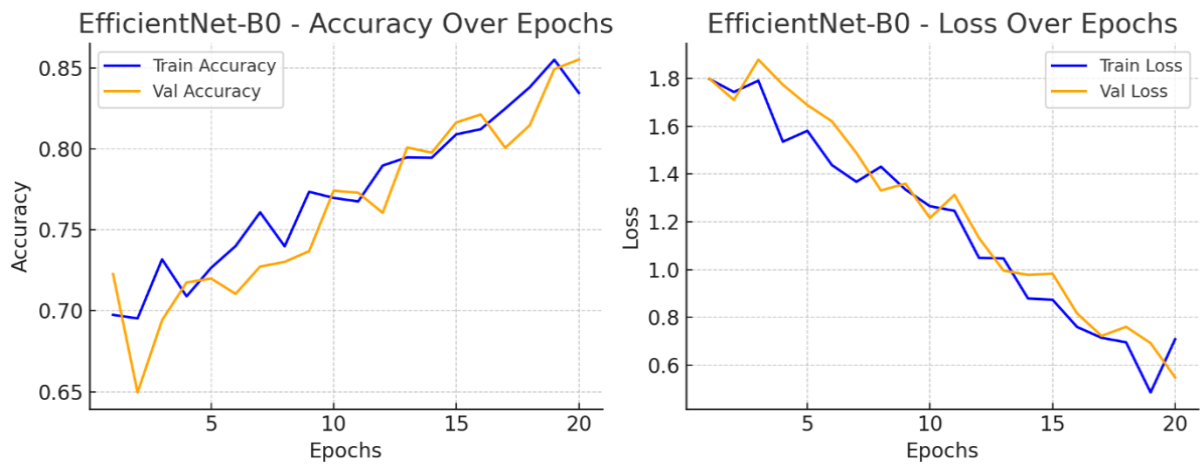

**Figure 7b**: Training and validation accuracy and loss trends for EfficientNet-B0, demonstrating smooth convergence and stable learning behavior.

### 6.1.3 SqueezeNet

**Lightweight Design and Performance Limitations**

SqueezeNet was evaluated for its compact design and low computational cost. The model achieved a training accuracy of 82.9% and a validation accuracy of 81.9%, but its classification performance was lower than both ResNet-18 and EfficientNet-B0. While SqueezeNet had the fastest inference time (11 ms), its lower accuracy and recall values made it less suitable for PerceptronCARE. The trade-off between speed and classification performance meant that although it was efficient for resource-limited environments, it was not optimal for real-time

DR screening. **Table 5** summarizes the performance of SqueezeNet, a highly lightweight model tested for its speed and compact design.

**Table 5: SqueezeNet Performance**

| Metric | Value |
| --- | --- |
| Training Accuracy | 82.9% |
| Validation Accuracy | 81.9% |
| Testing Accuracy | 81.7% |
| Training Loss | 0.421 |
| Validation Loss | 0.389 |
| Testing Loss | 0.415 |
| Model Size | 10.81 MB |
| Inference Time | 11 ms |
| FLOPs | 0.8 B |
| AUC Score | ~0.92 |

In the case of SqueezeNet, Figure 7c shows its training and validation curves, indicating rapid convergence and a relatively compact performance profile. Despite its lightweight architecture, SqueezeNet demonstrates strong learning capacity, supporting its feasibility for real-time screening tasks in constrained environments.

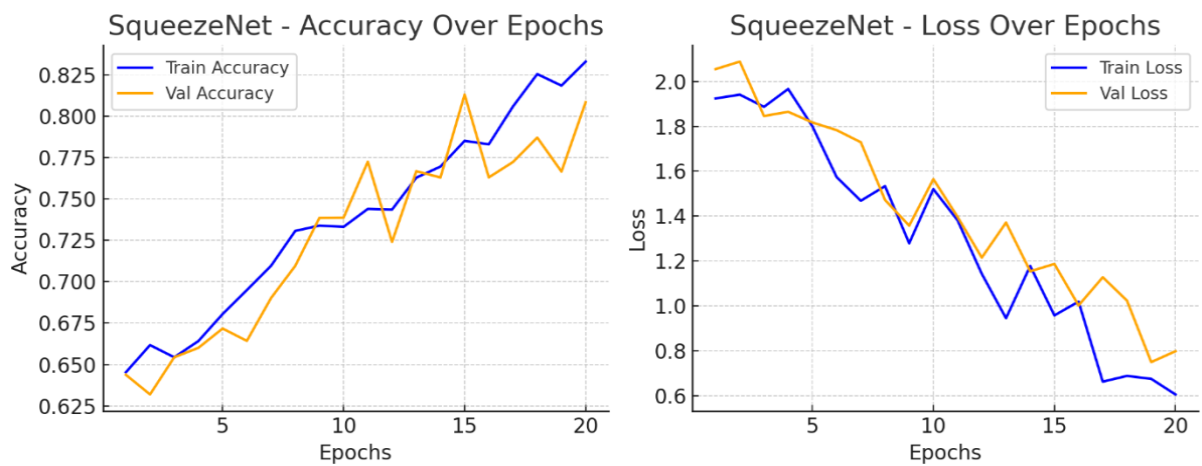

**Figure 7c**: Training and validation accuracy and loss trends for SqueezeNet, reflecting efficient learning within a lightweight architecture.

**6.2 Comparative Analysis of the Models and Justification for ResNet-18 Selection**

A comparative analysis of the models underscores their respective advantages and limitations. While EfficientNet-B0 and SqueezeNet demonstrated notable computational efficiency, ResNet-18 exhibited the highest accuracy and overall performance, thereby emerging as the most suitable model for integration into PerceptronCARE. Among the architectures tested,

ResNet-18 achieved an accuracy of 85.4% and an AUC score of 0.97, demonstrating superior classification reliability. The presence of residual connections enabled deeper feature extraction, which enhanced its ability to differentiate between varying severity levels of diabetic retinopathy.

Notwithstanding its greater depth compared to SqueezeNet, ResNet-18 maintained a balanced computational profile, with a model size of 8.18MB and an inference time of 18 milliseconds, ensuring its suitability for real-time screening. While EfficientNet-B0 exhibited competitive classification accuracy, its marginally lower performance and slightly larger computational footprint rendered it less favorable for integration. SqueezeNet, although designed for computational efficiency, demonstrated a notable drop in classification performance, making it a suboptimal choice for a diagnostic system requiring high precision and reliability.

Through the optimization of ResNet-18 via quantization, PerceptronCARE achieves real-time diabetic retinopathy classification without compromising diagnostic accuracy. This optimization ensures that the system remains computationally efficient while retaining high classification performance, making it well-suited for deployment in cloud-based hospital networks and mobile health applications. The ability to integrate ResNet-18 in a quantized format further enhances its scalability, allowing PerceptronCARE to support large-scale screening initiatives in diverse healthcare environments. The **Table 6** illustrates the summary performance of the models.

**Table 6: Summary of Model Performance**

| Model | Training Accuracy (%) | Validation Accuracy (%) | Testing Accuracy (%) | Model Size (MB) | Inference Time (ms) | FLOPs (B) | AUC Score | Precision (%) | Recall (%) | F1-Score (%) |
|---|---|---|---|---|---|---|---|---|---|---|
| ResNet-18 | 86.3 | 85.4 | 85.4 | 8.18 | 18 | 1.8 | ~0.97 | 85.2 | 85.0 | 85.1 |
| EfficientNet-B0 | 84.7 | 84.3 | 84.1 | 9.67 | 14 | 0.9 | ~0.95 | 84.1 | 83.9 | 84.0 |
| SqueezeNet | 82.9 | 81.9 | 81.7 | 10.81 | 11 | 0.8 | ~0.92 | 81.5 | 81.3 | 81.4 |

### 6.3 System Deployment and Scalability

PerceptronCARE follows a modular microservices-based architecture designed for scalable and efficient real-time diabetic retinopathy (DR) detection. The backend is built using Flask and FastAPI, providing a robust and lightweight framework for handling RESTful API requests and integrating deep learning-based automated diagnosis. In addition to these features, PerceptronCARE supports a practical and accessible image acquisition workflow where a smartphone can be directly attached to a standard ophthalmoscope. Through this setup, high-quality retinal fundus images can be captured using the smartphone and immediately uploaded to the PerceptronCARE platform for automated AI-based diagnosis. This flexibility enhances the platform's utility and deployment in low-resource settings and community outreach programs, making it a powerful tool for accessible and scalable diabetic retinopathy screening.

#### *6.3.1 Cloud Deployment and Scalability*

PerceptronCARE is deployed on Heroku, a cloud platform that supports seamless horizontal scaling to maintain stable performance under varying user loads. The system is containerized using Docker, ensuring portability, version control, and consistent behavior across diverse deployment environments. Kubernetes is used to orchestrate multi-instance deployments, automatically scaling inference services in response to real-time demand.

To efficiently handle high-throughput and concurrent processing of retinal fundus images, PerceptronCARE is built on Flask and employs asynchronous request handling combined with multi-threaded WSGI servers such as Gunicorn or uWSGI. This architecture supports WSGI-based concurrency, enabling the system to manage simultaneous API requests with low latency. Load balancing and distributed model serving further enhance responsiveness, allowing the platform to process over 100,000 retinal images efficiently, even during peak usage. This deployment strategy makes PerceptronCARE highly suitable for both clinical and teleophthalmology applications, where reliability, scalability, and rapid inference are essential.

#### *6.3.2 Real-Time Inference Optimization*

The DR detection model is deployed using TensorFlow Serving, an optimized framework for handling multiple inference requests with minimal delay. To improve real-time performance, the system leverages:
- Model quantization, reducing the computational complexity of deep learning inference while preserving diagnostic accuracy.

- GPU acceleration, allowing the model to process high-resolution fundus images efficiently.
- Optimized inference pipelines, ensuring that diagnostic results are delivered in real-time, enhancing usability in clinical workflows.

### 6.3.3 Database Management and Secure Data Storage

A structured database system is integrated for efficient data handling. PostgreSQL manages structured patient records, while MongoDB stores unstructured image metadata, ensuring seamless storage and retrieval of large medical datasets. This hybrid approach enhances both data integrity and query efficiency, supporting the scalability of PerceptronCARE across multiple healthcare facilities.

### 6.3.4 System Accessibility and Cross-Platform Support

PerceptronCARE is a web-based application, ensuring accessibility across multiple platforms, including Windows, macOS, Linux, and Android. Users can interact with the system via any web browser, eliminating the need for platform-specific installations. The cloud-based architecture enables remote access to DR screening services, making it suitable for both hospital networks and telemedicine frameworks.

### 6.3.5 Technical Novelty and Contribution

Unlike conventional DR detection systems, PerceptronCARE incorporates:

1. Quantization-based optimization, significantly reducing inference latency while maintaining diagnostic precision.
2. Scalable cloud deployment, ensuring adaptability to increased user demand.
3. Real-time telemedicine integration, allowing seamless connectivity with electronic health record (EHR) systems for automated patient diagnosis.
4. Edge and mobile compatibility, enabling low-power inference on mobile devices for decentralized DR screening.

By integrating these advancements, PerceptronCARE stands out as a novel AI-driven DR diagnosis platform, addressing scalability, computational efficiency, and clinical accessibility in a single unified system.

# 7 PRESENTING OUR APPLICATION: PERCEPTRONCARE DIABETIC RETINOPATHY APPLICATION

PerceptronCARE is a comprehensive teleophthalmology application designed to facilitate the early detection and diagnosis of diabetic retinopathy using deep learning. The application, built using Python Flask, provides a robust and user-friendly platform that supports three distinct user privileges: **User**, **Doctor**, and **Super Admin**. Below is a detailed overview of the functionality and features available to each user type, along with indications for where images would enhance the representation of the system's capabilities.

### A. User Platform

The user interface of PerceptronCARE is designed to be intuitive and user-friendly, enabling healthcare professionals to upload retinal images and receive diagnostic results with minimal effort. Upon launching the application, users are prompted to log in, ensuring secure access to patient data. The main dashboard provides options for uploading new images, viewing previous analyses, and generating reports as seen in the **Figure 8**.

*User Dashboard and Frontend*

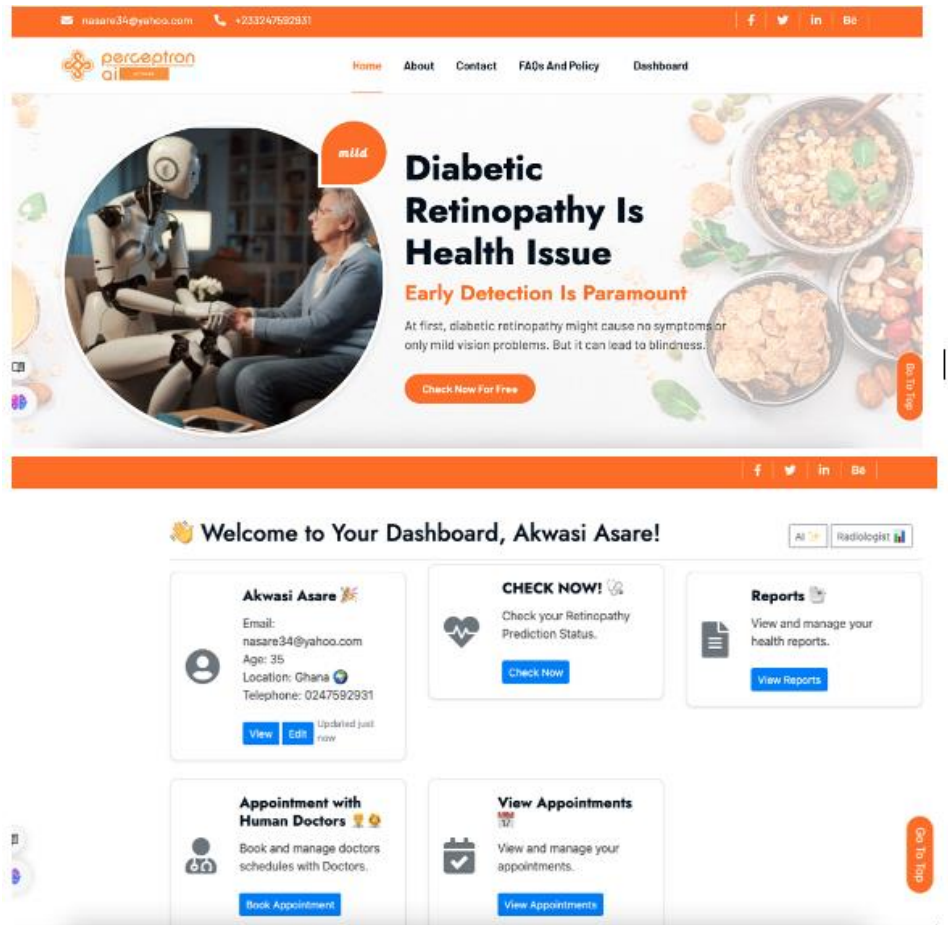

**Figure 8:** User dashboard showing options to upload images, check DR status, view history, and manage appointments.

### User Registration and Login

Users must register before gaining access to the system as depicted in the **Figure 9**. Registration involves creating an account with personal details, including email and password. After registration, the user can log in to access the platform's functionalities.

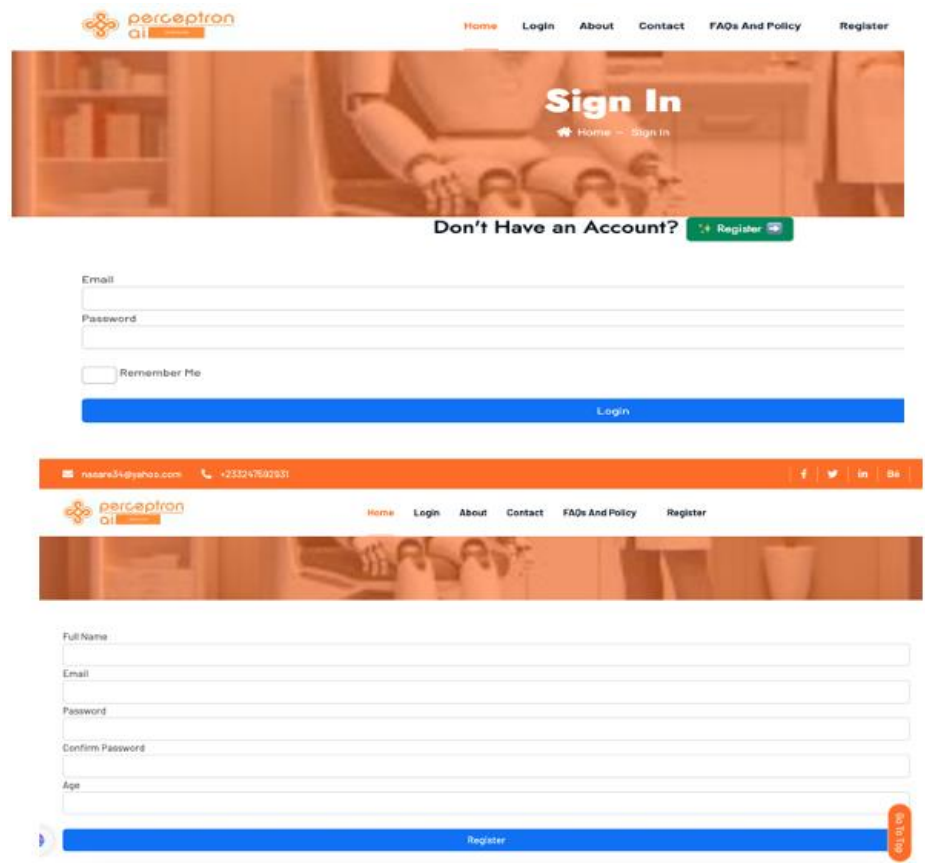

**Figure 9**: User registration and login interface for secure access.

**Uploading Retinal Images**

The system allows users to upload fundus images for the right and left eyes as depicted in the **figure 10**. It validates image types to ensure correct eye classification for diagnosis. Once logged in, users can upload retinal images specifically for the right and left eyes. The system is designed to ensure that only the correct type of image (right or left eye) is uploaded for analysis.

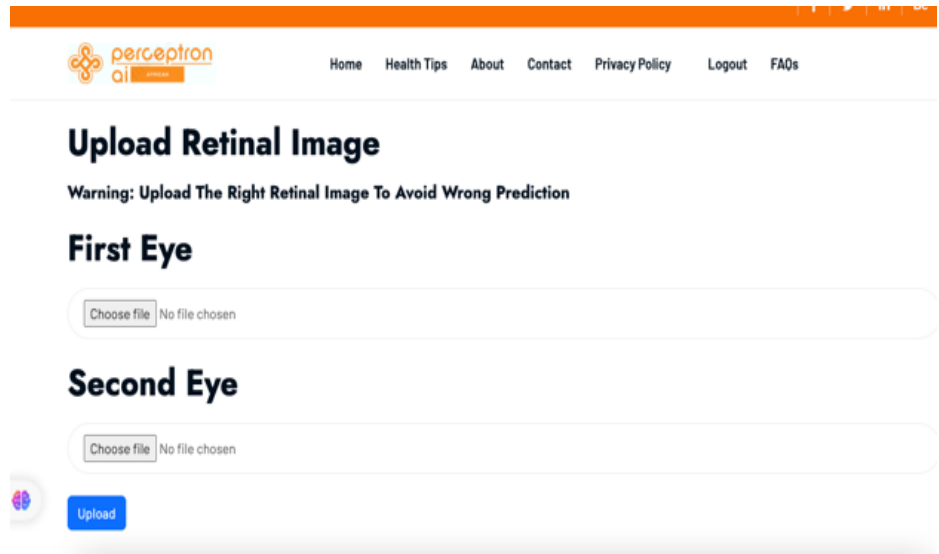

**Figure 10**: Upload interface for right and left eye images with validation prompts.

**Predicting Diabetic Retinopathy Status**

After uploading the retinal images, users can click the "Check" button to initiate the deep learning model's prediction process. The model analyzes the images and predicts the status of diabetic retinopathy. The core functionality of PerceptronCARE revolves around its ability to process retinal fundus images and provide a diagnosis of diabetic retinopathy severity. Users can upload images directly through the application interface, after which the images undergo preprocessing, including resizing, normalization, and enhancement. The processed images are then fed into the trained ResNet-18 model, which classifies them into different severity levels of diabetic retinopathy as seen in the **figure 11**.

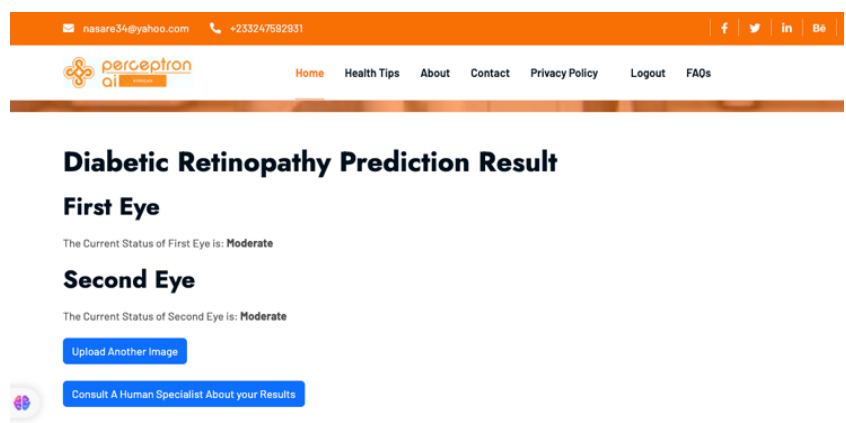

**Figure 11**: Prediction results interface showing severity level and model output.

**Viewing Prediction Results and Report History**

From the dashboard, users can view the prediction results and access a history of all previous reports, allowing them to track their diagnoses over time. As illustrated in the **figure 12**, All previous prediction results are saved and can be accessed from the dashboard, allowing users to monitor their DR status over time.

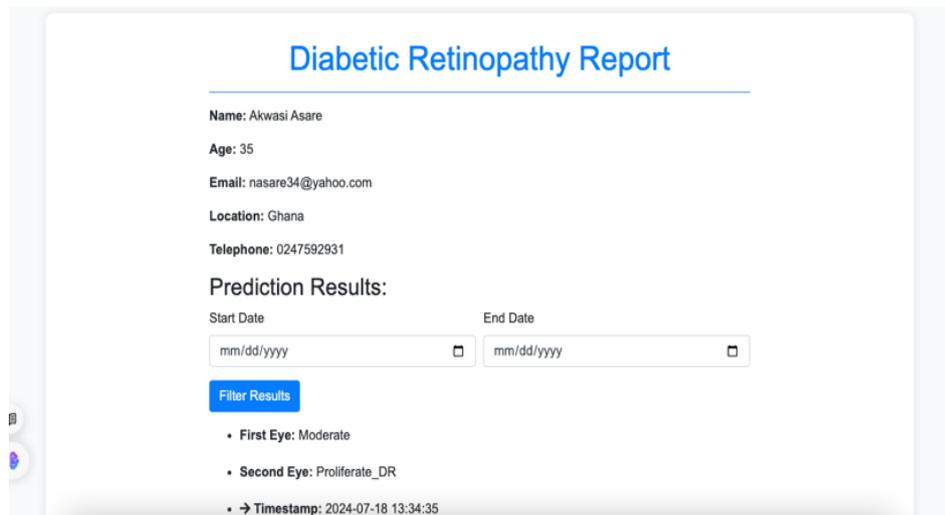

**Figure 12**: Report history page showing past DR prediction records for users.

**Booking and Canceling Appointments**

As seen in the **figure 13**. Users can book an appointment with a doctor directly from the application. Upon booking, the user receives an email confirmation. If necessary, users can also cancel appointments, which triggers a cancellation email. Confirmation or cancellation triggers automated email notifications.

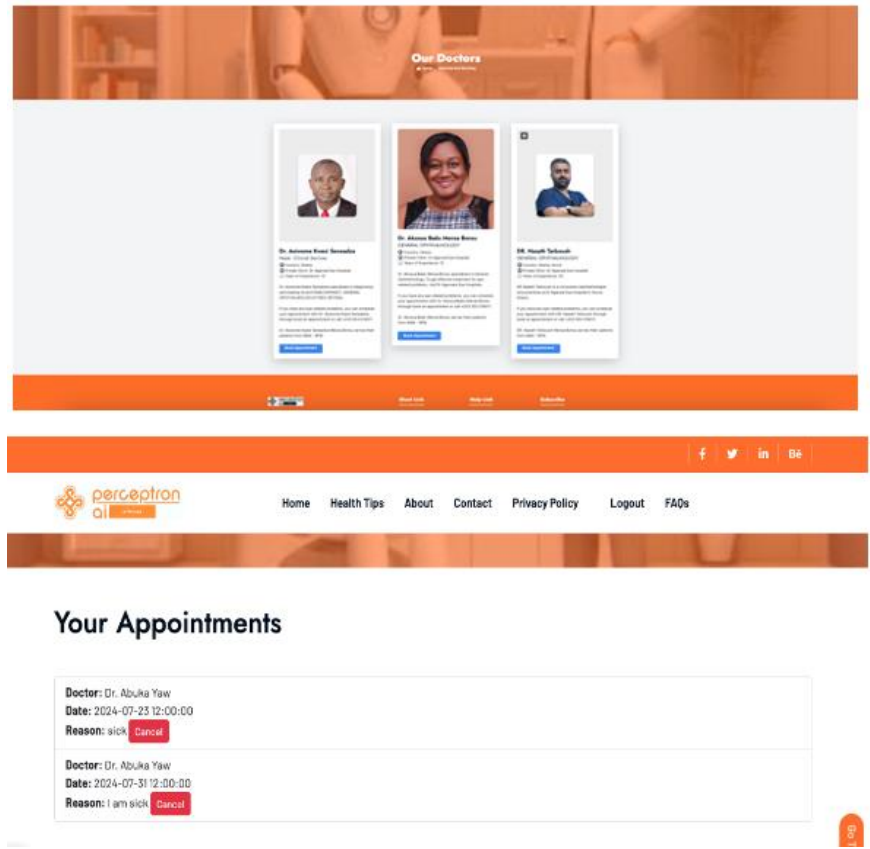

**Figure 13**: Appointment booking and cancellation interface with email notification confirmation.

**Generating and Downloading Reports**

Users have the option to generate a detailed report of their prediction results, which can be downloaded in PDF format as seen in the **figure 14**. This feature is particularly useful for keeping personal health records or sharing results with other healthcare providers.

**Figure 14**: Downloadable report generation module in PDF format for offline access.

**B. Doctor Platform**

**Doctor Registration and Login**

Doctors must register and log in to access their platform as seen in the **figure 15**. However, doctor registration requires special permission from the Super Admin, ensuring that only authorized medical professionals can access patient data.

**Figure 15**: Doctor registration interface with pending approval status.

**Managing Appointments and Viewing Patient Reports and Cases**

Doctors can view and manage appointments booked by users as illustrated in the **figure 16**. This includes receiving notifications of new appointments and the ability to cancel appointments if necessary. Like the user platform, doctors receive email confirmations for appointment bookings and cancellations. Doctors can access detailed reports and patients' histories. This allows them to review the diagnostic results and prepare for consultations.

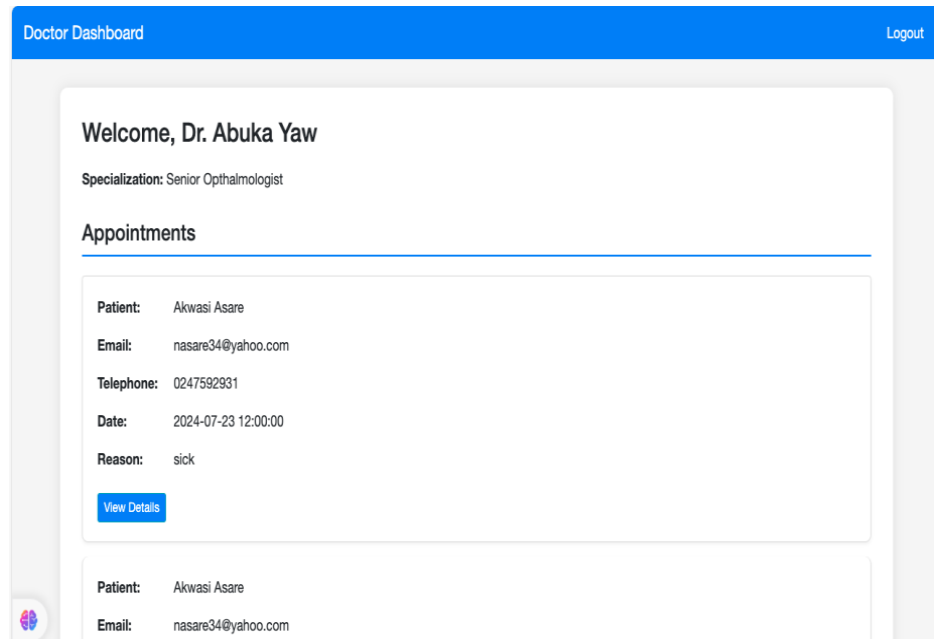

**Figure 16**: Doctor dashboard showing patient appointments and diagnostic report viewer.

**C. Super Admin Platform**

**Overseeing activities and Managing Users and Doctors**

The Super Admin has a comprehensive view of all activities within the application, including the ability to monitor user predictions and doctor appointments as depicted in the **figure 17**. This level of access ensures that the Super Admin can oversee the proper functioning of the application. The Super Admin can add or remove users and doctors from the system. This includes granting special permissions for doctor registrations and managing user access rights.

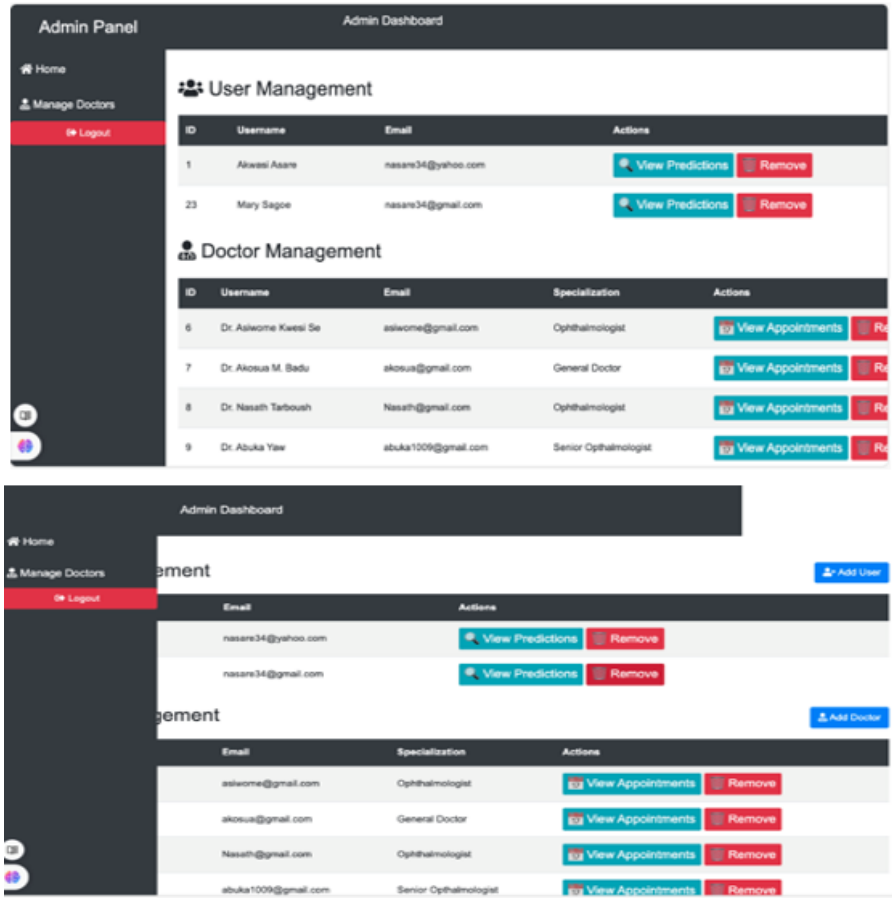

**Figure 17**: Super Admin control panel displaying system activity logs and user management tools.

**Cancelling Appointments**

As seen in the **figure 18**, Super Admins can cancel any scheduled appointments if required, ensuring administrative control and user protection and also, providing an additional layer of control over the scheduling process.

**Figure 18**: Appointment management panel for Super Admins, showing cancellation functionality.

# 8 CONCLUSION AND FUTURE WORK

This study presents PerceptronCARE, an AI-driven teleophthalmology platform designed for early diabetic retinopathy (DR) detection and classification. To ensure the system's effectiveness, three convolutional neural network (CNN) architectures—ResNet-18, EfficientNet-B0, and SqueezeNet—were evaluated based on classification accuracy, computational efficiency, and real-time feasibility. The ResNet-18 model was ultimately selected due to its superior balance between performance and efficiency. While EfficientNet-B0 offered a lower number of FLOPs (0.9B) and competitive accuracy (84.1%), it demonstrated slightly lower recall, which is crucial for minimizing false negatives in medical diagnoses. SqueezeNet, despite being the most lightweight with the fastest inference time (11ms), exhibited lower classification accuracy (81.7%), making it less reliable for clinical application.

ResNet-18 achieved the highest classification accuracy of 85.4%, with strong generalization capabilities attributed to its residual connections, which mitigate vanishing gradient issues and enhance feature extraction. After quantization, the model's memory footprint was significantly reduced, enabling fast and efficient inference suitable for both cloud-based and mobile deployment. The study also addressed dataset class imbalances by applying targeted augmentation techniques, improving the model's ability to detect all DR severity levels with higher sensitivity. In addition to its deep learning core, PerceptronCARE's architecture was designed for scalability, incorporating cloud-based deployment with GPU acceleration, Kubernetes orchestration for dynamic resource allocation, and a microservices framework to ensure reliable and concurrent model inference.

Moving forward, future enhancements will focus on further optimizing PerceptronCARE's diagnostic capabilities and expanding its applicability in ophthalmology. One key area of improvement is the integration of more advanced deep learning architectures, such as Vision Transformers (ViTs) or EfficientNet variants, which could further improve classification performance. Additionally, to enhance accessibility, the system will be optimized for mobile and edge computing environments using TensorFlow Lite or ONNX, enabling real-time DR screening on smartphones and portable medical devices. Another critical advancement will be the expansion of diagnostic capabilities beyond DR to include other ophthalmic conditions such as glaucoma, age-related macular degeneration (AMD), and hypertensive retinopathy, making PerceptronCARE a more comprehensive screening tool.

To improve the interpretability of model predictions and foster clinician trust, explainable AI (XAI) techniques such as Grad-CAM or SHAP visualizations will be integrated, providing transparent reasoning behind automated diagnoses. Additionally, efforts will be made to ensure interoperability with Electronic Health Record (EHR) systems, allowing seamless data exchange between PerceptronCARE and hospital infrastructures for automated referrals and long-term patient monitoring. Another future direction involves exploring federated learning approaches, enabling hospitals and medical institutions to collaboratively improve model performance while maintaining patient data privacy.

With these planned advancements, PerceptronCARE will continue evolving into a robust, scalable, and clinically reliable AI-driven teleophthalmology platform. Its deployment in both urban hospitals and remote, resource-limited settings will contribute to more accessible, early-stage DR screening, ultimately reducing preventable blindness and improving patient outcomes worldwide.

**Author Contributions:** The authors confirm their contribution to the paper as follows: study conceptualization, methodology, software, validation, formal analysis, writing – original draft, writing – review & editing, visualization**: Akwasi Asare & Isaac Baffour Senkyire**. Project administration, supervision, data curation, resources, validation: **Emmanuel Freeman.** Resources, writing – review & editing: **Akwasi Asare, Isaac Baffour Senkyire, Simon Hilary Ayinedenaba Aluze-Ele, Kelvin Kwao.**

**Data Availability Statement**

The dataset used in this study is publicly available on Kaggle as part of the **Diabetic Retinopathy Detection** competition, provided by EyePACS. It can be accessed at Diabetic Retinopathy Detection - Kaggle. Any preprocessing steps or modifications applied to the dataset in this study are available upon request from the corresponding author.

**Funding Statement**

This research did not receive any financial support from public, private, or non-profit funding agencies.

**Ethics Approval Statement**

This study does not involve human participants or animal subjects; therefore, ethical approval was not required. The dataset used consists of anonymized retinal images provided by

EyePACS and made publicly available through Kaggle, ensuring compliance with ethical and legal standards.

**Permission to Reproduce Material from Other Sources**

No copyrighted material from third-party sources has been reproduced in this study. All referenced materials are appropriately cited and used in accordance with fair use policies.

**Conflicts of Interest:** The authors declare that we have no conflicts of interest to report regarding the present study